\pdfoutput=1

\documentclass[11pt,a4paper]{article}
\usepackage[hyperref]{ranlp2023}
\usepackage{times}
\usepackage{latexsym}

\aclfinalcopy 

\usepackage{times}
\usepackage{latexsym}

\usepackage[T1]{fontenc}

\usepackage[T2A, T1]{fontenc}
\usepackage[utf8]{inputenc}
\usepackage[english]{babel}

\usepackage{tcolorbox}       
\usepackage{fancyvrb}
\usepackage{fvextra}

\usepackage{tablefootnote}

\usepackage{microtype}

\usepackage{enumitem}  

\usepackage{inconsolata}

\usepackage{graphicx}

\usepackage{multirow}

\title{\textsc{KoWit-24}: A Richly Annotated Dataset of Wordplay in News Headlines}

\author{
 \textbf{Alexander Baranov\textsuperscript{1}},
 \textbf{Anna Palatkina\textsuperscript{4}},
 \textbf{Yulia Makovka\textsuperscript{3}},
 \textbf{Pavel Braslavski\textsuperscript{1,2}}
\\
 \textsuperscript{1}HSE University 
 \textsuperscript{2}Ural Federal University
 \textsuperscript{3} Independent Researcher
  \\ \textsuperscript{4}Interdisciplinary Transformation University Austria
  \\
\texttt{ambaranov@hse.ru, anna.palatkina@it-u.at, yulya.makovka@yandex.ru, pbras@yandex.ru}
}

\begin{document}
\maketitle
\begin{abstract}
We present \textsc{KoWit-24}, a dataset with fine-grained annotation of wordplay in 2,700 Russian news headlines. 
\textsc{KoWit-24} annotations include the presence of wordplay, its type, wordplay anchors, and entities the wordplay refers to.
Unlike the majority of existing collections of \textit{canned} jokes, \textsc{KoWit-24} provides wordplay \textit{contexts}~-- each headline is accompanied by the news lead and summary. The most common type of wordplay in the dataset is the transformation of collocations, idioms, and named entities~-- the mechanism that has been underrepresented in previous humor datasets. 
Our experiments with five LLMs show that there is ample room for improvement in wordplay detection and interpretation tasks. 
The dataset and evaluation scripts are available at \url{https://github.com/Humor-Research/KoWit-24}.
\end{abstract}

\section{Introduction}

\textit{Wordplay} refers to creative language use that often purposely violates the linguistic norms and aims to draw attention, entertain, and amuse the reader. This umbrella term incorporates various techniques, such as punning, spoonerism, oxymoron, portmanteau, and their combinations. Even most modern advanced LLMs fail to recognize humor and struggle with generating original jokes~\cite{ismayilzada2024creativity}. 
At the same time, sense of humor 
is a desirable trait of conversational agents~\cite{shin2023influence}.



\begin{figure}
    \centering
    \includegraphics[width=0.4\textwidth]{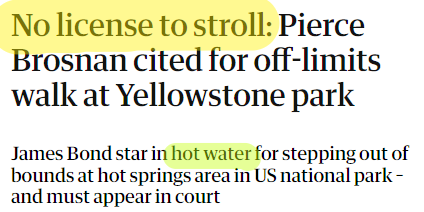}
    \setlength{\belowcaptionskip}{-10pt}
    \caption{Wordplay example from The Guardian. The part highlighted in yellow refers to \textit{Licence to Kill}, a concept popularized in the James Bond universe and the eponymous film, while the phrase in green allows both idiomatic and literal readings in this context. Source: \url{https://bit.ly/wpbrosnan}}
    \label{fig:guardian}
\end{figure}

A play on words is quite frequent in news~\cite{partington2009linguistic,monsefi2016wordplay}; see an example in Figure~\ref{fig:guardian}. In this paper, we present \textsc{KommersantWit (KoWit-24)}, a collection of headlines from the Russian business daily Kommersant that is known for its distinctive ironic style. The total size of the dataset is 2,700 headlines, about half of which are annotated as containing wordplay. Each wordplay-bearing headline is assigned up to two wordplay mechanisms from a set of eight 
and has an annotated \textit{anchor} (wordplay-triggering word or phrase). In addition, we provide a word, phrase, or entity the wordplay makes reference to along with a Wikipedia/Wiktionary link, if possible.
Importantly, wordplay examples in \textsc{KoWit-24} are contextualized: each headline is accompanied by a short description of the news story (lead) and a summary. 

\textsc{KoWit-24} has several features that distinguish it from other humor datasets:
1)~associated contexts, 2)~a large proportion of transformation-based wordplay examples underrepresented in the previous datasets, 3)~non-English content, 4)~multi-level annotation, and 5)~composition: items with and without wordplay come from the same source.
 
We conducted wordplay detection and interpretation experiments based on \textsc{KoWit-24} using a representative set of five LLMs. The results show that there is room for improvements even for GPT-4o, a definitive leader in both tasks. 
  

\section{Related Work}

In their pioneering paper, \citet{mihalcea2005making} presented a dataset containing 16k one-liners collected online and an equal number of non-humorous sentences.
Since then, several similar datasets have been released, including those that use \textit{reddit} as a source for humorous texts~\cite{yang2015humor,chen2018humor,weller-seppi-2020-rjokes,tang2023naughtyformer}. 
An alternative approach involves human editing: \citet{west2019reverse} designed an online game in which participants had to edit satirical headlines from \textit{TheOnion} to make them unfunny, while \citet{hossain2019president,hossain2020stimulating} explored the opposite direction: volunteers and crowd workers had to make news headlines funny with minimal editing. 
Several SemEval shared tasks have produced new datasets and sparked broader interest in computational humor~\cite{potash-etal-2017-semeval,hossain-etal-2020-semeval,meaney-etal-2021-semeval}. \citet{baranov-etal-2023-told} provide in-depth analysis of existing humor datasets.

While the majority of the datasets contain binary labels or funniness scores, a few provide more detailed annotations. EnglishPuns~\cite{miller-etal-2017-semeval} contains annotations of pun type 
and punning words along with their WordNet senses. 
\citet{zhang-etal-2019-telling} annotated a collection of Chinese jokes with keywords, character roles, place, humor category, and funniness score. 
EnglishPuns also became the basis for the ExPUN dataset~\cite{sun-etal-2022-expunations}, which additionally contains understandability,  offensiveness, and funniness scores, as well as keywords important for understanding the joke and natural language explanations. 

Most humor-related datasets are in English, but there are also datasets for Italian~\cite{buscaldi2007}, Spanish~\cite{castro2018}, and Portuguese~\cite{inacio-etal-2024-puntuguese}. The Russian FUN dataset~\cite{blinov-etal-2019-large} contains more than 150k funny short texts collected online and the same number of non-humorous forum posts. JOKER~\cite{ermakova2023joker} is a rare example of a bilingual collection: it extends EnglishPuns with French translations.

A study by~\citet{xu-etal-2024-good} is close to ours: they evaluate pun detection, explanation, and generation abilities of LLMs using English ExPUN dataset. 
\section{Methodology}

\subsection{Data Collection}

Kommersant is a Russian news outlet with both print and web editions.\footnote{\url{https://www.kommersant.ru/about} (in Russian)} Founded in 1990, the newspaper is one of the main Russian business dailies.  
Since its inception, Kommersant has developed its own distinctive ironic and playful style, which is best reflected in its headlines ~\cite{khazanov2023petrovich,chernyshova2021language,tymbay2024kommesant}.
  

%

We collected data from Kommersant via its RSS feed\footnote{\url{https://www.kommersant.ru/RSS/news.xml}; Kommersant grants permission to use its materials, provided that no more than 30\% of the original article is used, the text remains unaltered and an attribution is given,
see \url{https://www.kommersant.ru/copyright} (in Russian). The collected data complies with these requirements.} during the period from Jan 2021 to Dec 2023. Each data item corresponds to an article on the website and has the following fields: URL, category (World news, Business, etc.), 
headline, lead, summary, timestamp, and an optional image link.  

\subsection{Wordplay Definition}

Wordplay is a multifaceted and somewhat ambiguous concept. As the theoretical foundation for our work, we adopted the conceptual framework introduced and applied to the analysis of wordplay in a large collection of British news headlines by ~\citet{partington2009linguistic}.  
\citeauthor{partington2009linguistic} defines \textbf{two interpretations} associated with the text as the main characteristics of wordplay. In addition, these two meanings must be somewhat \textit{opposed}, and the wordplay must be deliberately constructed. \citeauthor{partington2009linguistic} distinguishes two main wordplay mechanisms: 1) \textit{relexicalization} and 2) \textit{reworking/reconstruction}. The former corresponds to traditional puns, where two meanings arise from either lexical ambiguity (homonymic puns) or phonetic ambiguity (homophonic puns). In the case of reworking/reconstruction, wordplay is based on the modification (\textit{reworking}) of a known phrase; its effect lies in the interplay of the meanings of the present phrase and the original one that the hearer/reader \textit{reconstructs}. 
\citeauthor{partington2009linguistic} points out that wordplay often involves different kinds of pre-constructed \textit{phrases}, such as proverbs, quotations, idioms, common collocations, film and book titles, etc. Note that this wordplay definition differs, for example, from those of~\citet{monsefi2016wordplay} and \citet{brugman2023humor}. In these studies linguistic devices such as personification, metaphor, metonymy, etc. in news headlines are attributed to wordplay.

\subsection{Data Annotation}

\renewcommand{\arraystretch}{1.5}
\begin{table*}[h!]
  \centering
  \footnotesize
  \begin{tabular}{p{0.1\textwidth}|p{0.45\textwidth}|p{0.4\textwidth}}
 
    \multicolumn{1}{c|}{\textbf{WP type}} & 
    \multicolumn{1}{c|}{\textbf{Original/transliterated headline and translated lead}} & \multicolumn{1}{c}{\textbf{Literal translation and interpretation}}\\ \hline
    
    Polysemy & \fontencoding{T2A}\selectfont{«Волгу» не могут заставить течь быстрее}   \newline Volgu ne mogut zastavit' tech' bystree \newline The speed limit on the M7 federal highway in the Moscow region remains unchanged & \textit{``Volga'' cannot be forced to flow faster.} Volga can both refer to the Volga river and federal highway ``Volga''.\\
    Homonymy & \fontencoding{T2A}\selectfont{Туризм подрастерял Шарм} \newline Turizm podrasteryal Sharm  \newline Operators adjust their Egyptian programs & \textit{Tourism has lost its charm.} Russian \textit{Sharm} can also refer to a shortened form of \textit{Sharm El Sheikh}, a holiday resort in Egypt. \\
    Homophony & \fontencoding{T2A}\selectfont{Из-под земли до стали} \newline Iz-pod zemli do stali  \newline The mineral extraction tax for metallurgical companies will be increased starting in 2022 & The headline sounds like an idiom \textit{Iz-pod zemli dostali}, literally \textit{Got out from under the ground}, equivalent to English \textit{Left no stone unturned}, but the spelling of the last word (\textit{dostali} $\rightarrow$ \textit{do stali}) allows for a different reading: \textit{From under the ground to the \underline{steel}}.\\
    Collocation & \fontencoding{T2A}\selectfont{Особо бумажные персоны} \newline Osobo bumazhnye persony  \newline How private investors are reshaping Russia’s IPO market & \textit{Very paper persons} refers to \textit{Very important persons}; Russian original/substitute words (\textit{vazhnye/bumazhnye}) rhyme.\\
    Idiom & \fontencoding{T2A}\selectfont{Код накликал} / Kod naklikal \newline Why and for whom open-source software matters in Russia & \textit{Code clicked} sounds in Russian similar to \textit{Cat cried}, an idiom describing a very small amount of something. \\
    Reference & \fontencoding{T2A}\selectfont{Миссия сократима} / Missiya sokratima \newline How Russia could respond to the expulsion of its diplomats from NATO’s Brussels mission & \textit{Mission reducible} refers to the \textit{Mission: Impossible} film series. \\ 
    Nonce word & \fontencoding{T2A}\selectfont{От запчастного к общему} \newline Ot zapchastnogo k obshchemu \newline  Car deficit sends Russian sales into reverse & \textit{From sparepartish to a whole}: the headline refers to the induction principle \textit{from a part to a whole} with a neologism adjective derived from the noun \textit{zapchast'} (\textit{spare part}).\\ 
    Oxymoron & \fontencoding{T2A}\selectfont{Новый премьер Израиля начал со старого} \newline Noviy premier Izrailya nachal so starogo \newline Naftali Bennett heads to Washington & \textit{\underline{New} Israeli PM started with \underline{old}} (tricks). \\ 

  \end{tabular}
  \caption{\label{tab:wordplay_examples}
    \fontencoding{T2A}\selectfont{Original wordplay examples along with interpretations.}
  }
\end{table*}
\renewcommand{\arraystretch}{1}

At the base of \textsc{KoWit-24} is the binary annotation of the wordplay presence. For the headlines with wordplay, we provide further annotations: 1)~wordplay type,
2)~\textit{anchors}, i.e. words or phrases that trigger the wordplay, 3)~\textit{anchor reference}, e.g. a similar-sounding word or original phrase the anchor refers to (note that there is no reference in case of homographic puns that are based on polysemous words), 
4)~for headlines that are modifications of a collocation, an idiom or refer to a popular entity (such as movie or book titles, catchphrases, etc.), we provide a corresponding link to Wiktionary or Wikipedia, if possible. 

The annotation was done using the Label Studio tool\footnote{\url{https://labelstud.io/}} by three authors of the paper, two of whom are professional linguists and one is a computer scientist; all three are Russian native speakers and have an extensive experience with NLP-related projects. Translated annotation guidelines can be found in the repository.

Three annotators labeled each element of the data in parallel, making notes on ambiguous cases that were later discussed. We compared the results, discussed discrepancies, and reconciled them in the annotation process. 
The average of three pairwise Cohen's kappas for the initial wordplay annotations \textit{before discussion} was 0.42, indicating the non-trivial nature of the task (two annotators with linguistic background showed better agreement with $\kappa = 0.58$).\footnote{Low agreement is typical for humor-related annotation.  
E.g., \citet{sun-etal-2022-expunations} reported Cohen's kappa values of 0.58 for joke definitions and 0.41 for joke ratings. Similar low agreement has been reported in sarcasm~\cite{oprea2020isarcasm} and insult~\cite{mathew2021hatexplain} detection studies.
} 
The majority of discrepancies were found in the Reference and Collocation categories. There was virtually no disagreement when annotating Oxymoron and Homonymy.
However, we hope that 
subsequent reconciliation of discrepancies ensures a high quality of the resulting annotation.
The overlapping annotation was particularly useful: the different cultural preferences and backgrounds of the annotators allowed to get a higher coverage, as not all interpretations are obvious and immediately understandable. 

Later, we assigned up to two mechanisms to the wordplay headlines identified in the first phase. The approach was mainly data-driven: we grouped the headlines based on the similarity of their wordplay mechanisms as we went through the collection, assigned labels, and occasionally re-annotated some items. The final list of the wordplay mechanisms used in the annotation is given in Table~\ref{tab:wordplay_examples} along with examples. The two main groups, \textit{Puns} and \textit{Transformations} (see Table~\ref{tab:wordplay_types}), correspond to the aforementioned mechanisms of \citet{partington2009linguistic} --  \textit{relexicalization} and \textit{reworking/reconstruction}. Within puns, we distinguish homonymic puns based on either on \textit{polysemy} or \textit{homonymy}, as well as homophonic puns based on \textit{phonetic similarity}. We classify transformation-based puns by their source phrases -- common \textit{collocations}, \textit{idioms}, or \textit{references} -- named entities such as quotations, book or film titles, etc. \textit{Nonce words} (occasionalisms) and \textit{oxymorons} form a separate group.  



\begin{table}[!t]
    \centering
    \small
    \begin{tabular}{llrrr}
         & \textit{Wordplay type} & \textit{\#} & \textit{AAL} & \textit{Links} \\ \hline
         \multirow{3}{*}{\rotatebox{90}{Puns}} 
         & Polysemy & 190  & 1.51 & \\
         & Homonymy & 26  &  1.57 & \\ 
         & Phonetic similarity & 98 & 1.80 \\ \hline
         \multirow{3}{*}{\rotatebox{90}{Trans.}} & Collocation  & 423 &  2.64 & 126\\
         & Idiom      & 177 &  3.43 & 118\\
         & Reference  & 353 &  3.73 & 214 \\    \hline
         & Nonce word & 185 & 1.44 & \\
         & Oxymoron   & 48  &  2.02 & \\
    \end{tabular}
    \caption{Wordplay types, average anchor length (AAL) in words, and wiki links in \textsc{KoWit-24}. Three mechanisms at the top of the table correspond to traditional \textit{puns}. Three mechanisms in the middle are based on \textit{transformations} of existing phrases. Note that some items are assigned two mechanisms, so the sum of the counts exceeds the number of headlines with wordplay in the dataset (1,340). The last column shows the number of wiki links for transformation-based types.}
    \label{tab:wordplay_types}
\end{table}

Finally, we annotated wordplay anchors, provided anchor references, and, 
if possible, added a link to the corresponding Wiktionary or Wikipedia page.\footnote{Wordplay examples in Figure~\ref{fig:guardian} would be annotated with \textit{Reference} and \textit{Polysemy} types, respectively. The highlighted spans would be annotated as wordplay anchors, with \textit{Licence to Kill} as the anchor reference accompanied by the corresponding Wikipedia link.} 
An example of a dataset record is shown in Figure~\ref{fig:structure_visualization} in the Appendix.

\subsection{Dataset Statistics and Analysis}

\begin{figure}
    \centering
    \small
    \includegraphics[width=.8\linewidth]{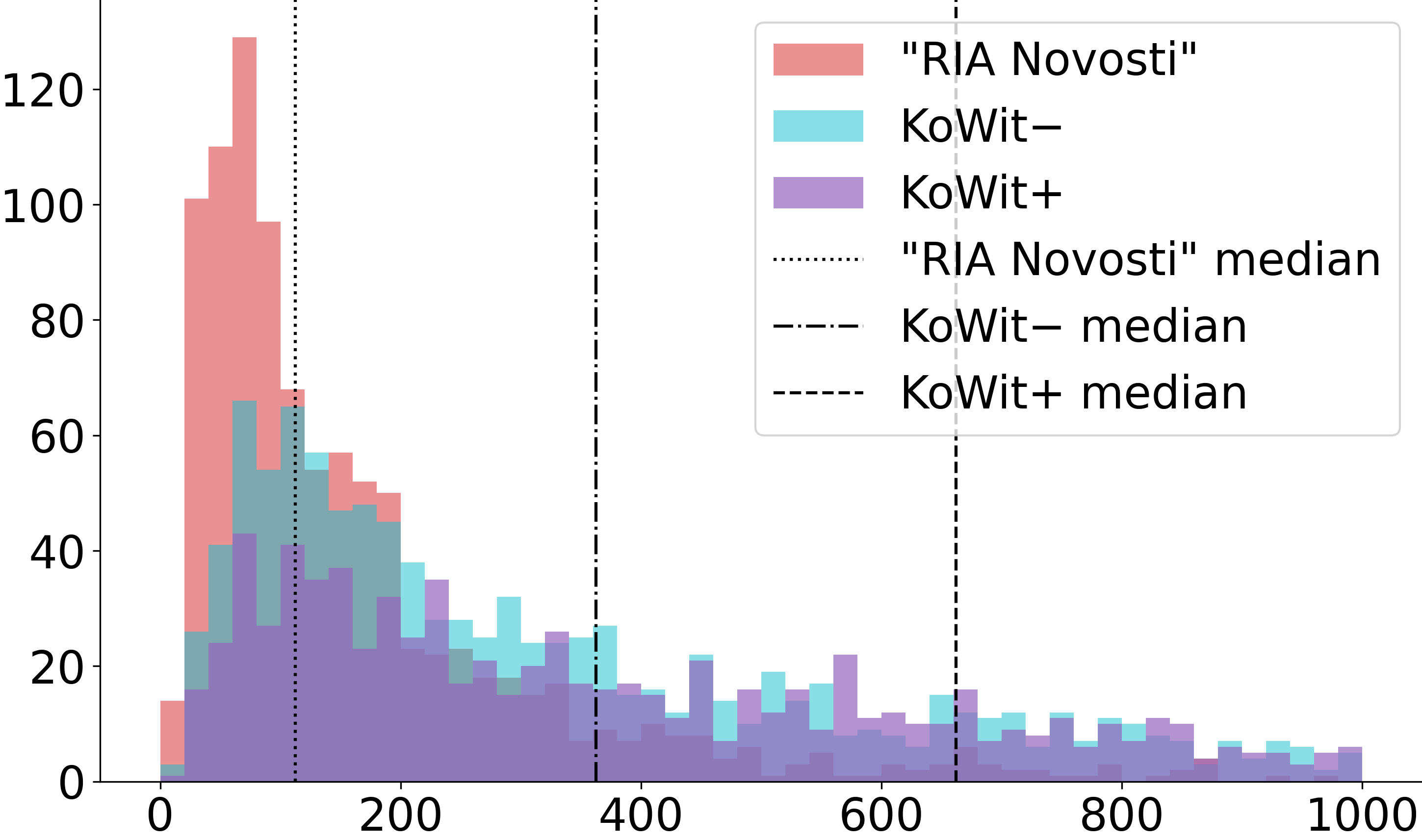}
    \caption{Perplexity distribution of the headlines in two (+/--) \textsc{KoWit-24} classes and RIA Novosti collection. Vertical lines correspond to the medians of the distributions (note that the histogram is truncated).}
    \label{fig:perplexity}
\end{figure}

In total, we annotated 2,700 headlines, of which 1,340 contained wordplay, so the dataset is almost perfectly balanced. It is interesting to note that the wordplay headlines are on average one word shorter than their counterparts (3.88 vs. 4.81 words).
To characterize \textsc{KoWit‑24}, we calculated perplexity of the headlines in both classes using ruGPT-3.5 13B model~\cite{rugpt13b} and juxtaposed them with the news headlines from the RIA Novosti dataset~\cite{gavrilov2019self}. 
As Figure~\ref{fig:perplexity} shows, Kommersant headlines have higher perplexity than the more reserved headlines of the state-owned agency RIA Novosti, and the \textsc{KoWit-24} headlines with wordplay deem even more `unusual' than their counterparts.

Distribution of headlines by wordplay type can be seen in Table~\ref{tab:wordplay_types}. The most frequent wordplay mechanism in our dataset appeared to be the modification of existing  well-known phrases~-- collocations, idiomatic expressions, or named entities. 
Notably, this type of wordplay is barely presented in previous humor datasets. As anticipated, the average anchor lengths are also higher in the transformation-based classes (average anchor length in the whole collection is 2.65 words). Our observations are in good agreement with the study by~\citet{partington2009linguistic}, who argues that the wordplay occurs mainly at the \textit{phrasal} level.


About half of all headlines with transformation-based wordplay are provided with Wikipedia (290) and Wiktionary (168) links pointing to descriptions of the source phrases/entity names. The presence of an article in one of the wikis can be seen as an indicator of the popularity of the original phrase/entity, which reduces the risk of subjective and spurious associations. 




\section{Experiments}

\begin{table*}[t!]
    \centering
    \small
\begin{tabular}{l|r|cc|cc|cc|cc}

         \textit{Wordplay type} & \textit{\#} & \multicolumn{2}{c|}{\textit{GigaChat Lite}} & \multicolumn{2}{c|}{\textit{GigaChat Max}} & \multicolumn{2}{c|}{\textit{YaGPT4}} & \multicolumn{2}{c}{\textit{GPT-4o}}\\
         & & simple & extended & simple & extended & simple & extended & simple & extended\\
         \hline
         Polysemy      & 168    & 0.56  & 0.74  & 0.57  & 0.57 & 0.05 & 0.23 & \textbf{0.88} & 0.86 \\
Homonymy               & 22     & 0.50   & 0.59  & 0.50   & 0.64 & 0.14 & 0.23 & 0.68 & \textbf{0.82} \\
Phonetic similarity    & 88     & 0.40   & 0.74  & 0.44  & 0.58 & 0.10  & 0.15 & 0.81 & \textbf{0.90} \\
Collocation            & 393    & 0.47  & 0.70   & 0.48  & 0.58 & 0.09 & 0.20  & 0.78 & \textbf{0.87} \\
Idiom                  & 164    & 0.50   & 0.74  & 0.49  & 0.65 & 0.12 & 0.38 & 0.87 & \textbf{0.96} \\
Reference              & 326    & 0.49  & 0.71  & 0.44  & 0.58 & 0.06 & 0.23 & 0.76 & \textbf{0.85} \\
Nonce word             & 166    & 0.52  & 0.81  & 0.45  & 0.60  & 0.21 & 0.27 & 0.87 & \textbf{0.96} \\
Oxymoron               & 34     & 0.68  & 0.79  & 0.68  & 0.71 & 0.21 & 0.41 & \textbf{0.85} & 0.82 \\
\hline
Total                    & 1,361   & 0.50  & 0.72  & 0.48  & 0.59 & 0.10 & 0.24 & 0.81 & \textbf{0.88} \\
        
    \end{tabular}    
    \caption{Recall on wordplay detection by type with simple/extended prompt (Mistral's all-zero scores are not shown).}
     \label{tab:scores_by_type}
\end{table*}

\begin{table*}[t!]
    \centering
    \small
\begin{tabular}{l|r|cc|cc|cc|cc|cc}

      \textit{Wordplay type} & \textit{\#} & \multicolumn{2}{c|}{\textit{GigaChat Lite}} & \multicolumn{2}{c|}{\textit{GigaChat Max}} & \multicolumn{2}{c|}{\textit{YaGPT4}} & \multicolumn{2}{c|}{\textit{Mistral}} & \multicolumn{2}{c}{\textit{GPT-4o}} \\
       &  & manual & auto & manual & auto & manual & auto & manual & auto & manual & auto \\
\hline
Polysemy & 12 & 0.17 & 0.33 & 0.33$\uparrow$ & 0.25 & 0.08 & 0.00 & 0.17 & 0.33 & \textbf{0.50} & 0.50 \\
Homonymy & 7 & 0.14 & 0.43 & 0.29 & 0.29 & 0.14 & 0.14 & 0.00 & 0.43 & \textbf{0.43}$\uparrow$ & 0.29 \\
Phonetic similarity & 85 & 0.11 & 0.25 & 0.28 & 0.39 & 0.11 & 0.21 & 0.15 & 0.32 & \textbf{0.52}$\uparrow$ & 0.51 \\
Collocation & 393 & 0.07 & 0.18 & 0.27 & 0.27 & 0.16 & 0.20 & 0.20 & 0.30 & \textbf{0.44}$\uparrow$ & 0.41 \\
Idiom & 164 & 0.15 & 0.18 & 0.37$\uparrow$ & 0.30 & 0.32$\uparrow$ & 0.28 & 0.34$\uparrow$ & 0.30 & \textbf{0.55}$\uparrow$ & 0.48 \\
Reference & 326 & 0.10 & 0.12 & 0.25$\uparrow$ & 0.23 & 0.20$\uparrow$ & 0.16 & 0.23 & 0.25 & \textbf{0.46}$\uparrow$ & 0.36 \\
Nonce word & 166 & 0.15 & 0.46 & 0.29 & 0.57 & 0.28 & 0.43 & 0.28 & 0.57 & \textbf{0.61} & 0.69 \\
Oxymoron & 6 & 0.67 & 0.67 & 0.50 & 0.50 & 0.33 & 0.33 & \textbf{0.83} & 0.83 & 0.67$\uparrow$ & 0.50 \\
\hline
Total & 1,159 & 0.11 & 0.19 & 0.28 & 0.28 & 0.20 & 0.22 & 0.24 & 0.30 & \textbf{0.48}$\uparrow$ & 0.43  \\

\end{tabular}
\caption{Recall on wordplay interpretation by type; manual and automatic evaluation ($\uparrow$ marks cases, where manual score exceeds the automatic one).}
\label{tab:manual_auto_interpretatio_metrics}
\end{table*}

For the experiments, we allocated 200 records (100 from each class) for the development set, making sure that all wordplay types were represented. Thus, the test set contains 2,500 headlines (1,290 with and 1,310 without wordplay). Since the dataset is primarily intended for experiments with modern LLMs in few- or zero-shot mode, we didn't allocate a dedicated training set. With 2k+ test items, \textsc{KoWit-24} should ensure low variance in repeated runs.

We experimented with five LLMs and two tasks~-- 1)~wordplay detection and 2)~wordplay interpretation. 
The five LLMs are a representative mix of open/closed, medium-sized/large, and Russian-centric/multilingual models. Details about the models can be found in Table \ref{tab:additionally-llms-info} in the Appendix. 


When using LLMs, 
the temperature was set to 0.1 for the GPT-4o, GigaChat Lite, GigaChat Max and YandexGPT4 models, and to 0.3 for the Mistral NeMo model, as per the developers' recommendations. For the wordplay detection task, the maximum number of generated tokens was set to 128, and for the wordplay interpretation task -- to 2,048. For GPT-4o, we used model version \textit{gpt-4o-2024-08-06}, with knowledge up-to-date as of October 2023. The YandexGPT4 model version is specified by its release date, and we used version 23.10.2024. The GigaChat Max version 26.10 was accessed through the API.

For the wordplay detection task, we employed two types of prompts in Russian: 1) a simple prompt asking whether the headline contains wordplay and 2) an extended prompt with definitions and two examples for each of eight wordplay types from the development set, see Table~\ref{tab:prompt_examples} in Appendix. When designing prompts, we adhered to the guidelines outlined in the OpenAI documentation under the \textit{Prompt engineering} section.\footnote{https://platform.openai.com/docs/guides/prompt-engineering/six-strategies-for-getting-better-results} 
In both cases, the LLM input included the headline and the lead. 


For the wordplay interpretation task, we used 1,033 headlines with annotated \textit{anchor references}, which are not present verbatim in the original headline and thus allow for a streamlined evaluation. The instruction and examples of wordplay were included in the prompt, similarly to the extended prompt in the detection task. In the automatic evaluation, we labeled the interpretation correct if we could match the lemmatized reference in the system's response (the approach is similar to automatic evaluation of pun explanation by~\citeauthor{xu-etal-2024-good}). 


\begin{table}[t!]
    \centering
    \small
        \begin{tabular}{l|cc|cc}  
            
             \textit{Model} & \multicolumn{2}{c|}{\textit{Detection, P / R}} & \multicolumn{2}{c}{\textit{Interpretation, R}}\\
                   & simple & extended & ~~manual & auto \\ \hline
             Giga Lite & 0.50 / 0.50 & 0.53 / 0.72 & ~~0.11  & 0.19 \\ 
             Giga Max & 0.62 / 0.48 & 0.68 / 0.59 & ~~0.28  & 0.28 \\
            YaGPT4  &  0.83 / 0.10  & 0.76 / 0.24 & ~~0.20  & 0.22 \\ 
            Mistral &  0.00 / 0.00  & 0.00 / 0.00 & ~~0.24 & 0.30 \\ 
            GPT-4o &  0.62 / 0.81  & 0.65 / 0.88 & ~~0.48 & 0.43 \\ 
        \end{tabular}
    \caption{Wordplay detection precision and recall using a simple/extended prompt and
    interpretation recall on headlines with anchor references based on manual/string matching scoring.}  \label{tab:results}
\end{table}

The results of the experiments are summarized in Table~\ref{tab:results}. GPT-4o demonstrates the strongest performance in both tasks, significantly outperforming the other four models. In the detection task, the extended prompt improves both precision and recall in three out of five models. 
The high precision of YandexGPT's detection  comes at the cost of low recall. Interestingly, Mistral returns only noes in the detection task, while it is quite competitive in the interpretation task. 
YandexGPT4 and GigaChat Max appeared to be  very strictly moderated: in the detection task with a simple prompt, they refused to give an answer and suggested changing the topic in 24.8\% and 15.4\% of cases, respectively.\footnote{The rejection rate is even higher for more straightforward RIA Novosti headlines -- 34.4\% and 27.4\%, suggesting that Aesopian language can partially overcome strict moderation.} 

Looking at the recall of wordplay recognition by type (Table~\ref{tab:scores_by_type}), we cannot conclude that some mechanisms are more challenging for all LLMs, but there are small variations within the results of each LLM. The extended prompt improves recall in almost all cases, sometimes significantly. However, the quality of GPT-4o's recognition of oxymorons and polysemy-based wordplay deteriorates slightly. The extended prompt doesn't change GigaChat Max's recall of polysemy-based wordplay. GPT-4o's recall of idiom-based wordplay and nonce words reaches 0.96 -- the best among all types. It can be assumed that non-dictionary occasionalisms are very different from the rest of the vocabulary, and idiom-based wordplay is easily recognized due to the frequency and stability of idiomatic expressions it refers to.

Interpretation scores are expectedly lower than detection scores. Again, idiom-based wordplay and nonce words seem to be slightly easier for interpretation; see Table~\ref{tab:manual_auto_interpretatio_metrics}.  Although not perfect, automatic evaluation seems to be a viable and efficient option in the interpretation task. Lower manual scores are largely due to hallucinations~-- the models often generate invented phrases that resemble the correct ones.\footnote{Similarly, when tasked with interpreting RIA Novosti headlines that actually contain no wordplay, LLMs often identify a polysemous word in the text and base their explanation on it.} In general, automatic evaluation inflates the scores compared to manual checking, but sometimes the opposite occurs, most notably in case of GPT-4o.
We carefully examined these cases and found that GPT-4o returns spelling variants or references that are slightly different from the canonical ones. 
These cannot be captured by straightforward string matching, but are considered correct by manual evaluation.  



The obtained results for both tasks are much lower than LLMs' recognition and explanation scores on English puns~\cite{xu-etal-2024-good}, though they cannot be directly compared.

\section{Conclusion}

In this paper we presented \textsc{KoWit-24}, a dataset of richly annotated wordplay in Russian news headlines. 
We demonstrated how the dataset can be used for wordplay detection and interpretation tasks. The provided multi-level annotation not only contributes to detailed linguistic analysis, but also enables automatic evaluation, which is a significant advantage for NLG tasks. Experiments with five models, which well reflect the variety of available LLMs, show that even advanced models such as GPT-4o face significant challenges in fully understanding and interpreting wordplay in Russian. We expect that the dataset can be used for other tasks as well. For example, previous studies suggest that rich annotation of jokes can improve humor generation~\cite{zhang2020let,sun-etal-2022-expunations,xu-etal-2024-good}.

We have made the dataset, evaluation scripts, and all code to reproduce the experiments available.
We hope that \textsc{KoWit-24} will facilitate  research in the field of multilingual computational humor.



\paragraph{Acknowledgments.} We thank Vladislav Kirienko who developed first version of the script for collecting Kommersant data. PB is partially supported by the Ural Mathematical Center (project \#075-02-2025-1719/1). The experiments were partially conducted on HPC facilities at HSE University. Access to YandexGPT4 was possible through Yandex Cloud grant. We thank anonymous reviewers for their insightful comments.


\section{Limitations}
There are several limitations to study wordplay in headlines. 
First, the annotation process is inherently subjective, as the identification of wordplay may vary depending on individual interpretation, educational background, etc. However, we hope that the implemented procedure ensures a high quality of the resulting annotation.
Second, the specific editorial style of Kommersant introduces bias, as the outlet is known for its particular style and language, which may not be representative of broader journalistic practices.
Third, because we lack visibility into the models’ training data, we cannot confirm that the news items in our dataset were entirely absent from those training sets.
In addition, the experiments used only five LLMs and did not involve extensive prompt engineering, meaning that the reported results can potentially be improved with more effective prompts and the use of different LLMs. We have not assessed variations in LLMs' outputs upon repeated attempts.



\section{Ethical considerations}
Our dataset reveals instances of wordplay even in the headlines of articles about sensitive topics such as diseases, death, and war, which some readers may find unacceptable. We will add a warning to the published dataset.  
\bibliographystyle{acl_natbib_ranlp}

\begin{thebibliography}{40}
\expandafter\ifx\csname natexlab\endcsname\relax\def\natexlab#1{#1}\fi

\bibitem[{Achiam et~al.(2023)Achiam, Adler, Agarwal, Ahmad, Akkaya, Aleman, Almeida, Altenschmidt, Altman, Anadkat et~al.}]{achiam2023gpt}
Josh Achiam, Steven Adler, Sandhini Agarwal, Lama Ahmad, Ilge Akkaya, Florencia~Leoni Aleman, Diogo Almeida, Janko Altenschmidt, Sam Altman, Shyamal Anadkat, et~al. 2023.
\newblock Gpt-4 technical report.
\newblock \emph{arXiv preprint arXiv:2303.08774}.

\bibitem[{Baranov et~al.(2023)Baranov, Kniazhevsky, and Braslavski}]{baranov-etal-2023-told}
Alexander Baranov, Vladimir Kniazhevsky, and Pavel Braslavski. 2023.
\newblock \href {https://doi.org/10.18653/v1/2023.emnlp-main.845} {You told me that joke twice: A systematic investigation of transferability and robustness of humor detection models}.
\newblock In \emph{Proceedings of the 2023 Conference on Empirical Methods in Natural Language Processing}, pages 13701--13715, Singapore. Association for Computational Linguistics.

\bibitem[{Blinov et~al.(2019)Blinov, Bolotova-Baranova, and Braslavski}]{blinov-etal-2019-large}
Vladislav Blinov, Valeria Bolotova-Baranova, and Pavel Braslavski. 2019.
\newblock \href {https://doi.org/10.18653/v1/P19-1394} {Large dataset and language model fun-tuning for humor recognition}.
\newblock In \emph{Proceedings of the 57th Annual Meeting of the Association for Computational Linguistics}, pages 4027--4032, Florence, Italy. Association for Computational Linguistics.

\bibitem[{Brugman et~al.(2023)Brugman, Burgers, Beukeboom, and Konijn}]{brugman2023humor}
Britta~C. Brugman, Christian Burgers, Camiel~J. Beukeboom, and Elly~A. Konijn. 2023.
\newblock \href {https://doi.org/10.1080/15205436.2022.2144747} {Humor in satirical news headlines: Analyzing humor form and content, and their relations with audience engagement}.
\newblock \emph{Mass Communication and Society}, 26(6):963--990.

\bibitem[{Buscaldi and Rosso(2007)}]{buscaldi2007}
Davide Buscaldi and Paolo Rosso. 2007.
\newblock Some experiments in humour recognition using the italian wikiquote collection.
\newblock In \emph{Applications of Fuzzy Sets Theory}, pages 464--468. Springer Berlin Heidelberg.

\bibitem[{Castro et~al.(2018)Castro, Chiruzzo, Ros{\'a}, Garat, and Moncecchi}]{castro2018}
Santiago Castro, Luis Chiruzzo, Aiala Ros{\'a}, Diego Garat, and Guillermo Moncecchi. 2018.
\newblock \href {https://aclanthology.org/W18-3502} {A crowd-annotated {S}panish corpus for humor analysis}.
\newblock In \emph{Proceedings of the Sixth International Workshop on Natural Language Processing for Social Media}, pages 7--11, Melbourne, Australia. Association for Computational Linguistics.

\bibitem[{Chen and Soo(2018)}]{chen2018humor}
Peng-Yu Chen and Von-Wun Soo. 2018.
\newblock \href {https://doi.org/10.18653/v1/N18-2018} {Humor recognition using deep learning}.
\newblock In \emph{Proceedings of the 2018 Conference of the North {A}merican Chapter of the Association for Computational Linguistics: Human Language Technologies, Volume 2 (Short Papers)}, pages 113--117, New Orleans, Louisiana. Association for Computational Linguistics.

\bibitem[{Chernyshova(2021)}]{chernyshova2021language}
Tatyana Chernyshova. 2021.
\newblock Language mechanisms of building the ironic texts and ways of their linguistic research (linguistic pragmatic aspect).
\newblock \emph{The European Journal of Humour Research}, 9(1):57--73.

\bibitem[{Ermakova et~al.(2023)Ermakova, Bosser, Jatowt, and Miller}]{ermakova2023joker}
Liana Ermakova, Anne{-}Gwenn Bosser, Adam Jatowt, and Tristan Miller. 2023.
\newblock \href {https://doi.org/10.1145/3539618.3591885} {The {JOKER} corpus: English-french parallel data for multilingual wordplay recognition}.
\newblock In \emph{Proceedings of the 46th International {ACM} {SIGIR} Conference on Research and Development in Information Retrieval, {SIGIR} 2023, Taipei, Taiwan, July 23-27, 2023}, pages 2796--2806. {ACM}.

\bibitem[{Gavrilov et~al.(2019)Gavrilov, Kalaidin, and Malykh}]{gavrilov2019self}
Daniil Gavrilov, Pavel Kalaidin, and Valentin Malykh. 2019.
\newblock \href {https://doi.org/10.1007/978-3-030-15719-7\_11} {Self-attentive model for headline generation}.
\newblock In \emph{Advances in Information Retrieval - 41st European Conference on {IR} Research, {ECIR} 2019, Cologne, Germany, April 14-18, 2019, Proceedings, Part {II}}, volume 11438 of \emph{Lecture Notes in Computer Science}, pages 87--93. Springer.

\bibitem[{GigaChat~team et~al.(2025)GigaChat~team, Kosarev, Leleytner, Shchuckin, Berezovskiy, Smirnov, Kozlov, Averkiev, Ivan, Proshunin et~al.}]{valentin2025gigachat}
Mamedov GigaChat~team, Valentin, Evgenii Kosarev, Gregory Leleytner, Ilya Shchuckin, Valeriy Berezovskiy, Daniil Smirnov, Dmitry Kozlov, Sergei Averkiev, Lukyanenko Ivan, Aleksandr Proshunin, et~al. 2025.
\newblock Gigachat family: Efficient russian language modeling through mixture of experts architecture.
\newblock \emph{arXiv preprint arXiv:2506.09440}.

\bibitem[{Hendrycks et~al.(2021)Hendrycks, Burns, Basart, Zou, Mazeika, Song, and Steinhardt}]{hendrycks2021measuring}
Dan Hendrycks, Collin Burns, Steven Basart, Andy Zou, Mantas Mazeika, Dawn Song, and Jacob Steinhardt. 2021.
\newblock \href {https://openreview.net/forum?id=d7KBjmI3GmQ} {Measuring massive multitask language understanding}.
\newblock In \emph{9th International Conference on Learning Representations, {ICLR} 2021, Virtual Event, Austria, May 3-7, 2021}.

\bibitem[{Hossain et~al.(2019)Hossain, Krumm, and Gamon}]{hossain2019president}
Nabil Hossain, John Krumm, and Michael Gamon. 2019.
\newblock \href {https://doi.org/10.18653/v1/N19-1012} {{``}president vows to cut {\textless}taxes{\textgreater} hair{''}: Dataset and analysis of creative text editing for humorous headlines}.
\newblock In \emph{Proceedings of the 2019 Conference of the North {A}merican Chapter of the Association for Computational Linguistics: Human Language Technologies, Volume 1 (Long and Short Papers)}, pages 133--142, Minneapolis, Minnesota. Association for Computational Linguistics.

\bibitem[{Hossain et~al.(2020{\natexlab{a}})Hossain, Krumm, Gamon, and Kautz}]{hossain-etal-2020-semeval}
Nabil Hossain, John Krumm, Michael Gamon, and Henry Kautz. 2020{\natexlab{a}}.
\newblock \href {https://doi.org/10.18653/v1/2020.semeval-1.98} {{S}em{E}val-2020 task 7: Assessing humor in edited news headlines}.
\newblock In \emph{Proceedings of the Fourteenth Workshop on Semantic Evaluation}, pages 746--758, Barcelona (online). International Committee for Computational Linguistics.

\bibitem[{Hossain et~al.(2020{\natexlab{b}})Hossain, Krumm, Sajed, and Kautz}]{hossain2020stimulating}
Nabil Hossain, John Krumm, Tanvir Sajed, and Henry Kautz. 2020{\natexlab{b}}.
\newblock \href {https://doi.org/10.18653/v1/2020.acl-demos.28} {Stimulating creativity with {F}un{L}ines: A case study of humor generation in headlines}.
\newblock In \emph{Proceedings of the 58th Annual Meeting of the Association for Computational Linguistics: System Demonstrations}, pages 256--262, Online. Association for Computational Linguistics.

\bibitem[{Inacio et~al.(2024)Inacio, Wick-Pedro, Ramisch, Esp{\'i}rito~Santo, Chacon, Santos, Sousa, Anchi{\^e}ta, and Goncalo~Oliveira}]{inacio-etal-2024-puntuguese}
Marcio~Lima Inacio, Gabriela Wick-Pedro, Renata Ramisch, Lu{\'i}s Esp{\'i}rito~Santo, Xiomara S.~Q. Chacon, Roney Santos, Rog{\'e}rio Sousa, Rafael Anchi{\^e}ta, and Hugo Goncalo~Oliveira. 2024.
\newblock \href {https://aclanthology.org/2024.lrec-main.1167/} {Puntuguese: A corpus of puns in {P}ortuguese with micro-edits}.
\newblock In \emph{Proceedings of the 2024 Joint International Conference on Computational Linguistics, Language Resources and Evaluation (LREC-COLING 2024)}, pages 13332--13343, Torino, Italia. ELRA and ICCL.

\bibitem[{Ismayilzada et~al.(2024)Ismayilzada, Paul, Bosselut, and van~der Plas}]{ismayilzada2024creativity}
Mete Ismayilzada, Debjit Paul, Antoine Bosselut, and Lonneke van~der Plas. 2024.
\newblock Creativity in ai: Progresses and challenges.
\newblock \emph{arXiv preprint arXiv:2410.17218}.

\bibitem[{Khazanov(2023)}]{khazanov2023petrovich}
Pavel Khazanov. 2023.
\newblock \href {https://doi.org/https://doi.org/10.1111/russ.12477} {A petrovich inside of every new russian: The disciplinary regime of the capitalist “vanguard group” at 1990s kommersant}.
\newblock \emph{The Russian Review}, 82(3):470--485.

\bibitem[{Mathew et~al.(2021)Mathew, Saha, Yimam, Biemann, Goyal, and Mukherjee}]{mathew2021hatexplain}
Binny Mathew, Punyajoy Saha, Seid~Muhie Yimam, Chris Biemann, Pawan Goyal, and Animesh Mukherjee. 2021.
\newblock Hatexplain: A benchmark dataset for explainable hate speech detection.
\newblock In \emph{Proceedings of the AAAI conference on artificial intelligence}, volume~35, pages 14867--14875.

\bibitem[{Meaney et~al.(2021)Meaney, Wilson, Chiruzzo, Lopez, and Magdy}]{meaney-etal-2021-semeval}
J.~A. Meaney, Steven Wilson, Luis Chiruzzo, Adam Lopez, and Walid Magdy. 2021.
\newblock \href {https://doi.org/10.18653/v1/2021.semeval-1.9} {{S}em{E}val 2021 task 7: {H}a{H}ackathon, detecting and rating humor and offense}.
\newblock In \emph{Proceedings of the 15th International Workshop on Semantic Evaluation (SemEval-2021)}, pages 105--119, Online. Association for Computational Linguistics.

\bibitem[{Mihalcea and Strapparava(2005)}]{mihalcea2005making}
Rada Mihalcea and Carlo Strapparava. 2005.
\newblock \href {https://aclanthology.org/H05-1067} {Making computers laugh: Investigations in automatic humor recognition}.
\newblock In \emph{Proceedings of Human Language Technology Conference and Conference on Empirical Methods in Natural Language Processing}, pages 531--538, Vancouver, British Columbia, Canada. Association for Computational Linguistics.

\bibitem[{Miller et~al.(2017)Miller, Hempelmann, and Gurevych}]{miller-etal-2017-semeval}
Tristan Miller, Christian Hempelmann, and Iryna Gurevych. 2017.
\newblock \href {https://doi.org/10.18653/v1/S17-2005} {{S}em{E}val-2017 task 7: Detection and interpretation of {E}nglish puns}.
\newblock In \emph{Proceedings of the 11th International Workshop on Semantic Evaluation ({S}em{E}val-2017)}, pages 58--68, Vancouver, Canada. Association for Computational Linguistics.

\bibitem[{{Mistral AI team}(2024)}]{mistralblogpost}
{Mistral AI team}. 2024.
\newblock Mistral nemo.
\newblock \url{https://mistral.ai/news/mistral-nemo/}, Accessed 15.09.2024.

\bibitem[{Monsefi and Sepora(2016)}]{monsefi2016wordplay}
Roya Monsefi and Tengku Sepora. 2016.
\newblock \href {https://www.proquest.com/scholarly-journals/wordplay-english-online-news-headlines/docview/2188087108/se-2} {Wordplay in english online news headlines}.
\newblock \emph{Advances in Language and Literary Studies}, 7(2):68--75.

\bibitem[{{NLP Core Team}(2023)}]{nlpcoreteam_mmlu_ru}
{NLP Core Team}. 2023.
\newblock {MMLU-RU}: Massive multitask language understanding (ru/en).
\newblock \url{https://huggingface.co/datasets/NLPCoreTeam/mmlu_ru}.
\newblock Accessed 24.05.2025.

\bibitem[{Oprea and Magdy(2020)}]{oprea2020isarcasm}
Silviu Oprea and Walid Magdy. 2020.
\newblock isarcasm: A dataset of intended sarcasm.
\newblock In \emph{Proceedings of the 58th Annual Meeting of the Association for Computational Linguistics}, pages 1279--1289.

\bibitem[{Partington(2009)}]{partington2009linguistic}
Alan~Scott Partington. 2009.
\newblock \href {https://doi.org/https://doi.org/10.1016/j.pragma.2008.09.025} {A linguistic account of wordplay: The lexical grammar of punning}.
\newblock \emph{Journal of Pragmatics}, 41(9):1794--1809.

\bibitem[{Potash et~al.(2017)Potash, Romanov, and Rumshisky}]{potash-etal-2017-semeval}
Peter Potash, Alexey Romanov, and Anna Rumshisky. 2017.
\newblock \href {https://doi.org/10.18653/v1/S17-2004} {{S}em{E}val-2017 task 6: {\#}{H}ashtag{W}ars: Learning a sense of humor}.
\newblock In \emph{Proceedings of the 11th International Workshop on Semantic Evaluation ({S}em{E}val-2017)}, pages 49--57, Vancouver, Canada. Association for Computational Linguistics.

\bibitem[{{SaluteDevices team}(2023)}]{rugpt13b}
{SaluteDevices team}. 2023.
\newblock rugpt-3.5 13b. technical report.
\newblock \url{https://habr.com/ru/companies/sberbank/articles/730108/}, Accessed 15.09.2024.

\bibitem[{Shin et~al.(2023)Shin, Bunosso, and Levine}]{shin2023influence}
Hyunju Shin, Isabella Bunosso, and Lindsay~R Levine. 2023.
\newblock The influence of chatbot humour on consumer evaluations of services.
\newblock \emph{International Journal of Consumer Studies}, 47(2):545--562.

\bibitem[{Sun et~al.(2022)Sun, Narayan-Chen, Oraby, Cervone, Chung, Huang, Liu, and Peng}]{sun-etal-2022-expunations}
Jiao Sun, Anjali Narayan-Chen, Shereen Oraby, Alessandra Cervone, Tagyoung Chung, Jing Huang, Yang Liu, and Nanyun Peng. 2022.
\newblock \href {https://doi.org/10.18653/v1/2022.emnlp-main.304} {{E}x{PUN}ations: Augmenting puns with keywords and explanations}.
\newblock In \emph{Proceedings of the 2022 Conference on Empirical Methods in Natural Language Processing}, pages 4590--4605, Abu Dhabi, United Arab Emirates. Association for Computational Linguistics.

\bibitem[{Tang et~al.(2023)Tang, Cai, and Wang}]{tang2023naughtyformer}
Leonard Tang, Alexander Cai, and Jason Wang. 2023.
\newblock \href {https://doi.org/10.1609/AAAI.V37I13.27034} {The naughtyformer: {A} transformer understands and moderates adult humor (student abstract)}.
\newblock In \emph{Thirty-Seventh {AAAI} Conference on Artificial Intelligence, {AAAI} 2023, Thirty-Fifth Conference on Innovative Applications of Artificial Intelligence, {IAAI} 2023, Thirteenth Symposium on Educational Advances in Artificial Intelligence, {EAAI} 2023, Washington, DC, USA, February 7-14, 2023}, pages 16348--16349. {AAAI} Press.

\bibitem[{Tymbay(2024)}]{tymbay2024kommesant}
Alexey Tymbay. 2024.
\newblock \href {https://doi.org/10.1177/17504813241236907} {Reading `between the lines': How implicit language helps liberal media survive in authoritarian regimes. the kommersant telegram posts case study}.
\newblock \emph{Discourse \& Communication}, 18(4):557--591.

\bibitem[{Weller and Seppi(2020)}]{weller-seppi-2020-rjokes}
Orion Weller and Kevin Seppi. 2020.
\newblock \href {https://aclanthology.org/2020.lrec-1.753} {The r{J}okes dataset: a large scale humor collection}.
\newblock In \emph{Proceedings of the Twelfth Language Resources and Evaluation Conference}, pages 6136--6141, Marseille, France. European Language Resources Association.

\bibitem[{West and Horvitz(2019)}]{west2019reverse}
Robert West and Eric Horvitz. 2019.
\newblock \href {https://doi.org/10.1609/AAAI.V33I01.33017265} {Reverse-engineering satire, or ``paper on computational humor accepted despite making serious advances'''}.
\newblock In \emph{The Thirty-Third {AAAI} Conference on Artificial Intelligence, {AAAI} 2019, The Thirty-First Innovative Applications of Artificial Intelligence Conference, {IAAI} 2019}, pages 7265--7272. {AAAI} Press.

\bibitem[{Xu et~al.(2024)Xu, Yuan, Chen, and Yang}]{xu-etal-2024-good}
Zhijun Xu, Siyu Yuan, Lingjie Chen, and Deqing Yang. 2024.
\newblock \href {https://doi.org/10.18653/v1/2024.emnlp-main.657} {{\textquotedblleft}a good pun is its own reword{\textquotedblright}: Can large language models understand puns?}
\newblock In \emph{Proceedings of the 2024 Conference on Empirical Methods in Natural Language Processing}, pages 11766--11782, Miami, Florida, USA. Association for Computational Linguistics.

\bibitem[{{Yandex team}(2024)}]{yagptblogpost}
{Yandex team}. 2024.
\newblock Yandexgpt 4.
\newblock \url{https://ya.ru/ai/gpt-4}, Accessed 09.02.2025.

\bibitem[{Yang et~al.(2015)Yang, Lavie, Dyer, and Hovy}]{yang2015humor}
Diyi Yang, Alon Lavie, Chris Dyer, and Eduard Hovy. 2015.
\newblock \href {https://doi.org/10.18653/v1/D15-1284} {Humor recognition and humor anchor extraction}.
\newblock In \emph{Proceedings of the 2015 Conference on Empirical Methods in Natural Language Processing}, pages 2367--2376, Lisbon, Portugal. Association for Computational Linguistics.

\bibitem[{Zhang et~al.(2019)Zhang, Zhang, Liu, Lin, and Xia}]{zhang-etal-2019-telling}
Dongyu Zhang, Heting Zhang, Xikai Liu, Hongfei Lin, and Feng Xia. 2019.
\newblock \href {https://doi.org/10.18653/v1/D19-1673} {Telling the whole story: A manually annotated {C}hinese dataset for the analysis of humor in jokes}.
\newblock In \emph{Proceedings of the 2019 Conference on Empirical Methods in Natural Language Processing and the 9th International Joint Conference on Natural Language Processing (EMNLP-IJCNLP)}, pages 6402--6407, Hong Kong, China. Association for Computational Linguistics.

\bibitem[{Zhang et~al.(2020)Zhang, Liu, Lv, and Luo}]{zhang2020let}
Hang Zhang, Dayiheng Liu, Jiancheng Lv, and Cheng Luo. 2020.
\newblock Let's be humorous: Knowledge enhanced humor generation.
\newblock \emph{arXiv preprint arXiv:2004.13317}.

\end{thebibliography}

\appendix
\onecolumn

\fontencoding{T1}\selectfont

\section{Appendix}\label{sec:appendix}

\begin{table*}[!h]
  \centering
  \small
  \begin{tabular}{llrrr}
    \hline
    \textbf{Model}           & \textbf{Availability} & \textbf{xMMLU},\% & \textbf{Release date} & \textbf{N} \\ 
    \hline
    GigaChat Lite \cite{valentin2025gigachat} & Open & $58.38$\textsuperscript{$\alpha$} & 2024-12-13 & 20B \\
    GigaChat Max \cite{valentin2025gigachat} & Closed & $75.00$\textsuperscript{$\alpha$} & 2024-11-02 & ? \\
    YandexGPT4 \cite{yagptblogpost} & Closed & $65.00$\textsuperscript{$\beta$} &  2024-10-23 &  ? \\
    Mistral Nemo \cite{mistralblogpost} & Open & $59.20$\textsuperscript{$\alpha$} & 2024-07-18 & 12B \\
    GPT-4o \cite{achiam2023gpt} & Closed & $86.40$\textsuperscript{$\gamma$} & 2024-08-06 &  ? \\
    \hline
  \end{tabular}
  \caption{\label{tab:additionally-llms-info}
    Five LLMs in the study. The first three LLMs are Russian-centric. xMMLU scores refer to various versions of the original MMLU benchmark: $\alpha$~-- ruMMLU \cite{nlpcoreteam_mmlu_ru} is an open Russian version, $\beta$~-- yaMMLU, a proprietary Russian version by Yandex \cite{yagptblogpost}, $\gamma$ -- original English MMLU \cite{hendrycks2021measuring}; N -- total number of model parameters.
  }
\end{table*}

\fontencoding{T2A}\selectfont
~
\begin{figure}[h!]
    \centering
\input{json_listing}
    \caption{Dataset entry example (based on original JSON format, simplified for readability).}
    \label{fig:structure_visualization}
\end{figure}

\fontencoding{T1}\selectfont

\begin{table}[h!]
    \centering
    \small
\resizebox{\columnwidth}{!}{
\begin{tabular}{p{0.13\textwidth}|p{.38\textwidth}|p{.38\textwidth}}
          \multicolumn{1}{c|}{\textbf{Prompt type}} & 
          \multicolumn{1}{c|}{\textbf{Original prompt}} & 
          \multicolumn{1}{c}{\textbf{Translated prompt}} \\
         \hline
User prompt & \fontencoding{T2A}\selectfont{Заголовок новости: <headline>.\newline Cодержание новости: <lead>} & Headline: <headline>.\newline News content: <lead> \\ 
\hline
System prompt for wordplay detection & \fontencoding{T2A}\selectfont{Присутствует ли в заголовке новости игра слов?\newline Дай ответ с учетом содержания новости.\newline Отвечать можешь только `да', `нет' или `не знаю'.\newline <instruction>} & Does the news headline contain wordplay?\newline Give an answer considering the content of the news.\newline You can only answer `yes', `no' or `don't know'.\newline <instruction> \\
\hline
System prompt for wordplay interpretation & \fontencoding{T2A}\selectfont{Проанализируй заголовок новости в контексте ее содержания. Укажи, есть ли в заголовке игра слов. Если она есть, объясни смысл, использованные методы и связь с основным текстом. Если игры слов нет, то ответь "в заголовке нет игры слов". <instruction>} & Analyse the news headline in the context of its content. Identify whether there is wordplay in the headline. If there is, explain the meaning, the methods used and the relationship to the main text. If there is no wordplay, answer ‘there is no wordplay in the headline’. <instruction> \\
\hline
    \end{tabular}
    }
    \caption{User and system prompts  for wordplay detection and interpretation.}
     \label{tab:prompt_examples}
     
\end{table}

\newpage

\end{document}